\documentclass[letterpaper, 10 pt, conference]{ieeeconf}  

\IEEEoverridecommandlockouts                              

\usepackage{amsmath}
\usepackage{amssymb}
\usepackage{bm}
\usepackage{booktabs}
\usepackage{xcolor}
\usepackage{multirow}
\usepackage{graphicx} 
\usepackage{colortbl}
\usepackage{cuted}
\usepackage{capt-of}
\usepackage{threeparttable}
\usepackage{placeins}
\usepackage{float}

\definecolor{best}{RGB}{0, 255, 200} 
\definecolor{second}{RGB}{255, 200, 200} 

\usepackage[normalem]{ulem}

\title{\LARGE \bf
Spectral GS-SLAM: Observability-Aware, Degeneracy-Robust Tracking for Real-Time 3D Gaussian Splatting SLAM}

\author{
    Edward Beng Wai Tan$^{1}$, Siew-Kei Lam$^{1}$, and Dongshuo Zhang$^{*,1}$%
    \thanks{$^{1}$College of Computing and Data Science, Nanyang Technological University, Singapore.}%
    \thanks{$^{*}$Corresponding author.}%
}

\begin{document}

\maketitle
\begin{strip}
    \vspace{-2.5\baselineskip}   
    \centering
    \includegraphics[width=\textwidth]{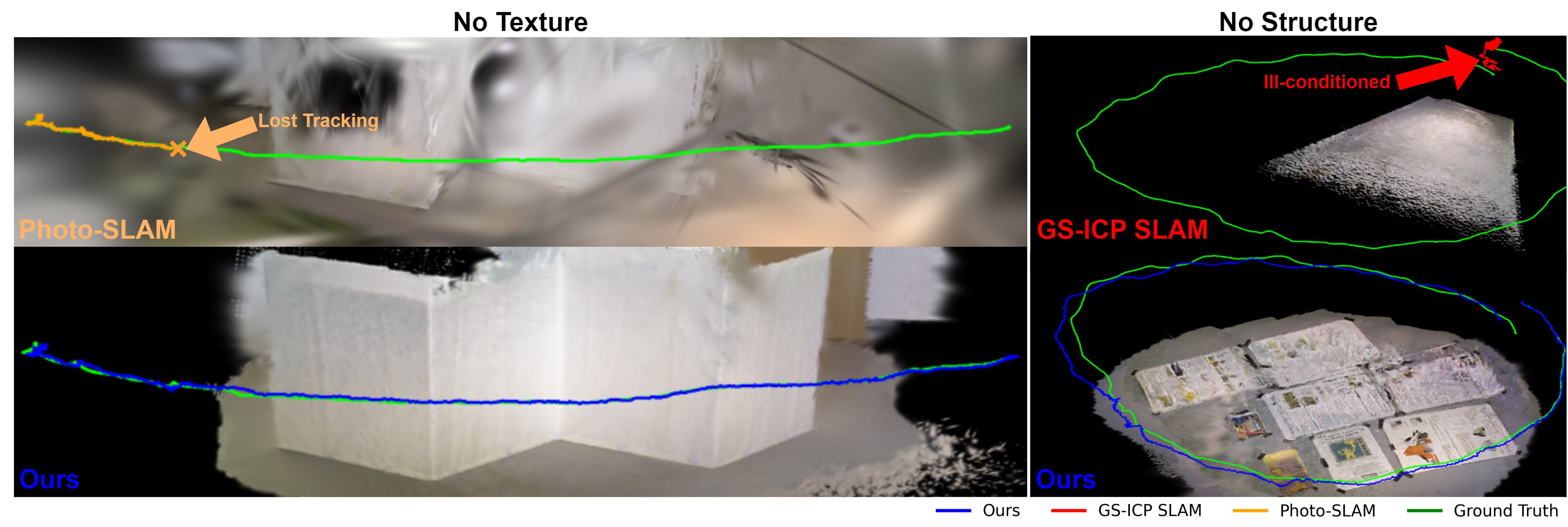}

    \captionof{figure}{In textureless scenes (left), feature-based methods such as \textcolor{orange}{Photo-SLAM} lose tracking due to insufficient visual correspondences. In planar, structureless environments (right), \textcolor{red}{GS-ICP SLAM} suffers from ill-conditioned optimization caused by dominant surface normals. In contrast, \textcolor{blue}{our} method maintains stable and accurate trajectories in both scenarios by adaptively compensating under-constrained directions through observability-aware information fusion.
    }
    \label{fig:hook}
\end{strip}
\thispagestyle{empty}
\pagestyle{empty}

\begin{abstract}

Recent 3DGS-SLAM systems enable real-time operation by leveraging conventional feature matching or ICP-based tracking, thereby avoiding the heavy dense photometric optimization used in earlier approaches. However, feature matching remains prone to failure in textureless environments, while ICP-based tracking struggles in structureless or geometrically degenerate scenes due to ill-conditioned optimization. To address this issue, we propose Spectral GS-SLAM, an efficient yet robust tracking framework that integrates ICP with complementary feature-based constraints. Our method mitigates numerical instability by adaptively compensating under-constrained directions in degenerate scenarios, without interfering with the shared Gaussian representation used for mapping. We further introduce a Gaussian-aware planarity weighting mechanism that exploits the intrinsic covariance structure of 3D Gaussians to characterize scene geometry and guide information fusion. Extensive evaluations on challenging TUM RGB-D sequences demonstrate that Spectral GS-SLAM achieves real-time performance (40.14 FPS) while maintaining consistent tracking in both structureless and featureless environments. The proposed method preserves trajectory integrity in degenerate scenes while maintaining competitive performance in non-adverse conditions.

\end{abstract}

\section{INTRODUCTION}

3D Gaussian Splatting (3DGS) has recently been integrated into Simultaneous Localization and Mapping (SLAM) systems, enabling photorealistic mapping with significantly higher rendering efficiency than earlier neural representations. While early 3DGS-SLAM frameworks~\cite{yan_gs-slam_2024, matsuki_gaussian_2024} achieved accurate tracking through dense photometric optimization, their high computational cost limits real-time deployment in robotic applications.

To enable real-time performance, recent 3DGS-SLAM systems adopt lightweight tracking modules, such as conventional feature matching or Generalized ICP (G-ICP)~\cite{segal_generalized-icp_2009}, while performing photometric Gaussian optimization in parallel. Although computationally efficient, these approaches inherit limitations of their tracking strategies: feature-based methods degrade in textureless environments, whereas ICP-based tracking becomes ill-conditioned in geometrically degenerate scenes.

Feature-based tracking methods such as Photo-SLAM~\cite{huang_photo-slam_2024} rely heavily on reliable detection and matching of ORB features~\cite{rublee_orb_2011}. However, they degrade significantly in featureless environments where stable keypoint correspondences cannot be established. Consequently, the tracking process becomes weakly constrained, often resulting in rapid drift or complete failure when observing large textureless regions (see Fig.~\ref{fig:hook}, left).

Unlike feature-based methods, ICP-based approaches such as GS-ICP SLAM~\cite{leonardis_rgbd_2025} are robust to texture deficiency, as they rely on geometric alignment rather than visual correspondences. However, they become unstable in geometrically degenerate scenes, where dominant surface normals induce near-singular covariance structures in the shared Gaussian representation. Under such degeneracy, the resulting Hessian matrix becomes severely ill-conditioned, reducing observability along certain motion directions and leading to unreliable pose estimation.

To address these challenges, we propose an observability-aware, degeneracy-robust tracking framework that effectively combines the complementary strengths of feature-based and ICP-based methods. Specifically, our system prioritizes ICP-based tracking and performs a spectral analysis of the Hessian matrix to identify under-constrained directions. This is guided by a novel ``Gaussian-aware" degeneracy detection that exploits the intrinsic covariance of 3D Gaussians to assess geometric reliability. Upon detecting ill-conditioned scenarios, the framework adaptively injects sparse feature-based constraints into the degenerate subspace. This targeted approach stabilizes optimization in either textureless or structureless environments while preserving the integrity of the shared Gaussian primitives.

Our contributions are summarized as follows:
\begin{enumerate}
    \item We propose a Gaussian-aware spectral degeneracy detection mechanism that characterizes geometric ill-conditioning via Hessian eigenvalue analysis and Gaussian normal statistics.
    \item We introduce an observability-aware subspace information fusion strategy that adaptively compensates under-constrained directions using lightweight feature-based priors.
    \item Our Spectral GS-SLAM achieves real-time performance (40.14 FPS) on the challenging TUM RGB-D sequences, while maintaining stable tracking under both featureless and structureless environments, and remaining robust in non-degenerate conditions.
\end{enumerate}

\section{RELATED WORK}

\textbf{Radiance Field Representations} were first integrated into SLAM systems for the goal of creating photorealistic maps. Initially, neural radiance fields \cite{mildenhall_nerf_2020} were used \cite{sucar_imap_2021, zhu_nice-slam_2022}, but suffered from the slow speed of volumetric rendering. Later works such as \cite{chung_orbeez-slam_2023, sandstrom_point-slam_2023, johari_eslam_2023}, though faster, still suffer from slow rendering speed.

\textbf{3D Gaussian Splatting (3DGS)}  \cite{kerbl_3d_2023} was later adopted by SLAM systems for its much faster rendering speed. However, early methods \cite{yan_gs-slam_2024, matsuki_gaussian_2024, keetha_splatam_2024} performed dense map-to-frame photometric optimization for pose estimation. The high computational cost meant that these methods were largely unable to run in real-time on most robotic platforms.

\textbf{Real-time 3DGS-SLAM} systems have recently been introduced, and by replacing dense photometric tracking with well-established and lightweight trackers, methods such as Photo-SLAM \cite{huang_photo-slam_2024} and GS-ICP SLAM \cite{leonardis_rgbd_2025} yield real-time performance, but inherit the limitations of their respective tracking algorithms. Feature matching methods like Photo-SLAM (based on ORB-SLAM3 \cite{campos_orb-slam3_2021}) fail to track in featureless environments, while ICP methods such as GS-ICP SLAM experience geometric degeneracy in structureless environments.

\textbf{Robust SLAM} systems have been proposed which fuse IMU measurements into the RGB \cite{leutenegger_keyframe-based_2015,qin_vins-mono_2018} or LiDAR \cite{shan_lio-sam_2020} sensors, and have recently been integrated into 3DGS-SLAM pipelines \cite{zhu_vigs-slam_2025, seiskari_gaussian_2024}, together with the heavy photometric tracker to solve the low texture and motion blur problem. Similarly, works like \cite{xiao_liv-gs_2025} use aligned LiDAR point clouds with 3DGS maps to overcome the resolution limitation of LiDAR. However, these methods have the limitation of requiring multi-modal sensing.

\textbf{LiDAR-based SLAM} works have extensively explored the ICP degeneracy problem, where \cite{zhang_degeneracy_2016} first proposed using eigenvalues to detect degenerate directions. Later works build on this by performing per-correspondence localizability checks \cite{tuna_x-icp_2024} and generalize the detection method to algorithms such as probabilistic ICP \cite{yue_lp-icp_2025}. These works are intended for LiDAR systems without awareness of visual features, and thus only address LiDAR-specific failure modes. In contrast, our method uses the rich information from a single RGB-D camera (visual features) to provide complementary information to the ICP-based tracker when the geometry fails, without the overhead of additional sensors.

\section{METHOD}

\begin{figure*}[t]
    \centering
    \includegraphics[width=0.98\textwidth]{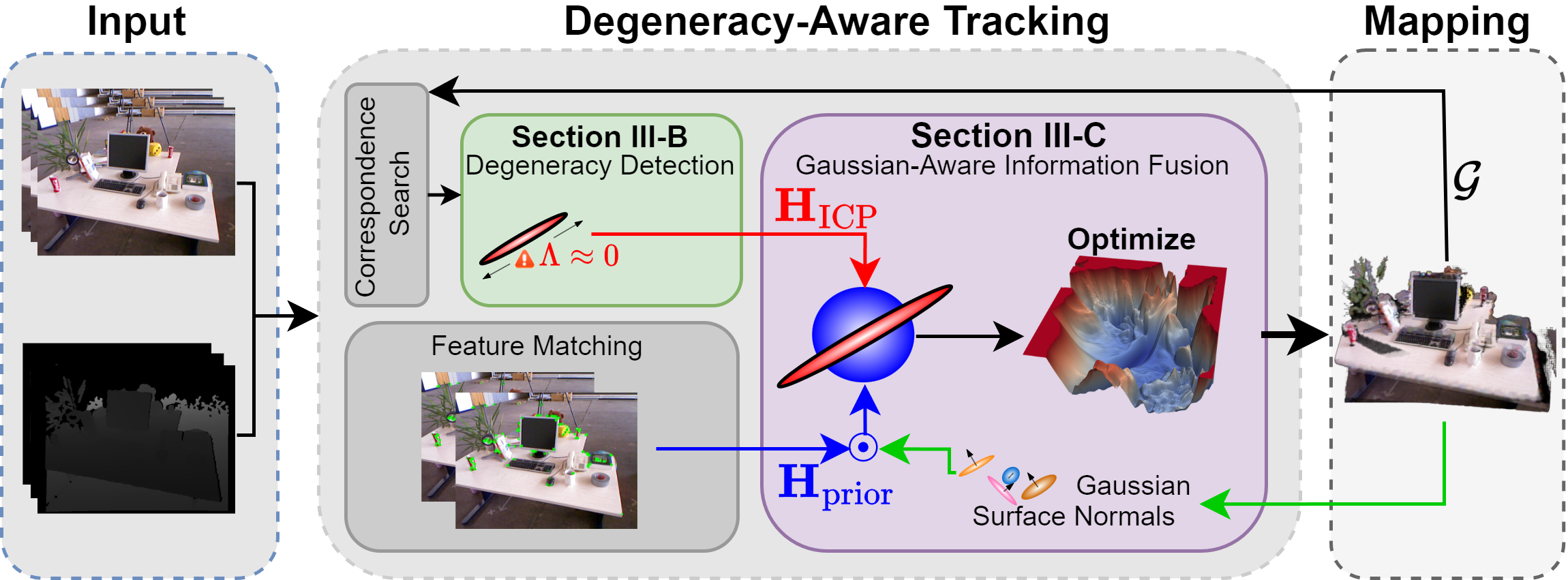}
    \vspace{5pt}
    \caption{\textbf{System Architecture.} Our pipeline consists of two main modules: (1) \textbf{Degeneracy Detection}, which detects if the optimization is ill-conditioned, and (2) \textbf{Gaussian-Aware Information Fusion}, which fuses the ill-conditioned ICP Hessian (\textcolor{red}{$\mathbf{H}_{\text{ICP}}$}) with complementary information (\textcolor{blue}{$\mathbf{H}_{\text{prior}}$}) from the feature-based prior, gated by the planarity of the visible Gaussian map.}
    \label{fig:system_overview}
\end{figure*}

Our proposed Spectral GS-SLAM is illustrated in Fig.~\ref{fig:system_overview}. Similar to GS-ICP SLAM \cite{leonardis_rgbd_2025}, we employ 3D Gaussian primitives $\mathcal{G}_i = \{\bm{\mu}_i, \bm{\Sigma}_i, \mathbf{c}_i, \alpha_i\}$ to provide a unified representation for both photorealistic mapping and G-ICP-based tracking. In this formulation, the Gaussian covariance $\bm\Sigma_i$ characterizes the local geometric structure, facilitating probabilistic registration.

Unlike point-to-point ICP, which minimizes the Euclidean distance between point correspondences, our tracking module performs distribution-to-distribution registration.

For each source point in the observation $\bm{\mu}^s_i$, the correspondence search finds the nearest neighbor $\bm{\mu}^t_j$ using the process:
\begin{equation}
    \pi (i) = \arg \min_j \ \lVert \bm{\mu}^s_i - \bm{\mu}^t_j \rVert _2
\end{equation}
Hence, the ICP residual is $\mathbf{d}_i = \bm{ \mu }^t_{\pi(i)} - \mathbf{T} \bm{\mu}^s_i$, and its uncertainty $\bm{\Sigma}_i = \bm{\Sigma}^t_{\pi(i)} + \mathbf{R} \bm{\Sigma}^s_i \mathbf{R}^\top$,
such that the optimal relative transformation $\mathbf{T} \in \mathrm{SE}(3)$ is found by minimizing the total Mahalanobis distance:
\begin{equation}
\mathbf{T}^* = \arg \min_{\mathbf{T}} \sum_{i} \mathbf{d}_i^\top \bm{\Sigma}_i^{-1} \mathbf{d}_i.
\label{eq:gicp_loss}
\end{equation}

This non-linear least squares problem is solved by finding the linearization of the residual $\mathbf{d}_i(\mathbf{T} \oplus \delta \boldsymbol{\xi})$, using a small pose increment $\delta \boldsymbol{\xi} \in \mathfrak{se}(3)$ in the tangent space: $\mathbf{d}_i(\delta \boldsymbol{\xi}) \approx \mathbf{d}_i(\mathbf{T}) + \mathbf{J}_i \delta \boldsymbol{\xi}$
where $\mathbf{J}_i = \frac{\partial \mathbf{d}_i}{\partial \delta \boldsymbol{\xi}}$ is the Jacobian matrix. Substituting this into the objective function in Eq. \eqref{eq:gicp_loss} and setting the derivative with respect to $\delta \boldsymbol{\xi}$ as zero yields the system's Hessian matrix:

\begin{equation}
\mathbf{H} = \sum_{i} \mathbf{J}_i^\top \bm{\Sigma}_i^{-1} \mathbf{J}_i  \in \mathbb{R}^{6 \times 6},
\end{equation}
which is used to calculate the update step.

\subsection{Problem Formulation}
Although the per-Gaussian covariance parameterization $\bm\Sigma$ is directly used by the ICP algorithm as described previously to quantify the spatial uncertainty, we find that during mapping, it is implicitly driven by the mapper towards flat, rank-2 Gaussians, to minimize the ${L}_1$ norm between the observation and the map over multiple views. Formally, given a surface normal $\mathbf{n} \in \mathbb{R}^3$, the scaling component $s_{\perp}$ along this normal direction is driven toward a minimal scale value $\epsilon$, while the lateral scales $s_{\parallel, 1}$ and $s_{\parallel, 2}$ remain significantly larger to maximize surface coverage. This is problematic as it results in a highly anisotropic scaling matrix $\mathbf{S} = \text{diag}(s_{\parallel, 1}, s_{\parallel, 2}, \epsilon)$. The resulting covariance $\bm{\Sigma} = \mathbf{R} \mathbf{S} \mathbf{S}^\top \mathbf{R}^\top$ possesses a highly skewed spectral distribution, where the eigenvalue corresponding to the normal direction, $\lambda_{\perp} = \epsilon^2$, is several orders of magnitude smaller than the lateral eigenvalues $\lambda_{\parallel, \{1,2\}}$. Consequently, the information matrix $\mathbf{\Lambda} = \bm{\Sigma}^{-1}$ exhibits a dominant eigenvalue $\lambda_{max} = 1/\epsilon^2$ in the direction of $\mathbf{n}$. Although this provides a sharp, well-constrained minimum for point-to-plane distances, the lateral information remains vanishingly small, i.e., $\mathbf{v}^\top \mathbf{\Lambda} \mathbf{v} \approx 0$ for any tangent vector $\mathbf{v} \perp \mathbf{n}$. 

For planar surfaces roughly perpendicular to the viewing ray, this collapses the noise model $\bm{\Sigma}$ in the viewing direction into a small value $\epsilon$ by averaging it over multiple views, which is important for strong directional constraints (consistent with the increased geometric stability observed with 2DGS \cite{huang_2d_2024} as primitives \cite{pak_g2s-icp_2025}), but renders the optimization ill-posed in geometrically degenerate scenes dominated by a single surface normal. In this case, numerical imbalance ill-conditions the tracking Hessian, creating unreliable and unstable pose estimation.

To solve this issue, while maintaining the mapper's optimization objective, and the strong directional constraints of ICP under geometrically stable scenes, we detect an information deficit during ICP optimization in Section~\ref{subsec:awareness}, then fuse an informative prior with complementary strengths into the under-constrained directions in Section~\ref{subsec:fusion}. 

\subsection{Degeneracy Detection}
\label{subsec:awareness}

We first aim to detect whether such a condition has occurred. Following the technique by \cite{zhang_degeneracy_2016}, we perform eigenvalue analysis of the positive semi-definite $\mathbf{H}$,
\begin{equation}
    \mathbf{H} = \mathbf{V} \operatorname{diag}(\lambda_1, ..., \lambda_6) \mathbf{V}^\top,
\end{equation}
where they identify that a particular eigenvector is ill-constrained if an eigenvalue is less than some heuristic threshold $\lambda_k < \tau$. We instead adopt a normalized condition-number metric $\frac{\lambda_{\text{max}}}{\lambda_k} > \tau_{\text{deg}}$ for reliable \textit{detection}. To determine \textit{how much} of the prior to fuse, we use an observability measure of the Gaussian normals' covariance, that utilizes the property of the mapper's optimization as described previously, to characterize the macroscopic scene geometry.

We leverage the existing Gaussian covariance parameterization, such that computing the Gaussian normals becomes an $O(1)$ operation (compared to point-based ICP where this would require a kd-tree search per point), and derive the unit vector $\mathbf{n}_i$ corresponding to $s_{\perp}$. Firstly, we compute the joint covariance $\mathbf{C}$ of the alpha-weighted Gaussian normals,
\begin{equation}
    \mathbf{C} = \frac{1}{\sum_{i=1}^N \alpha_i} \sum_{i=1}^N \alpha_i (\mathbf{n}_i - \bar{\mathbf{n}})(\mathbf{n}_i - \bar{\mathbf{n}})^\top
\end{equation}
where $\bar{\mathbf{n}}$ is the mean normal. We then perform the eigendecomposition $\mathbf{C} = \mathbf{Q}  \operatorname{diag}(\ell_1, \ell_2, \ell_3) \mathbf{Q}^\top$ with $\ell_1 \le \ell_2 \le \ell_3$.
The observation planarity is thus defined as
\begin{equation}
    \eta \propto \frac{\ell_3}{\ell_2 + \epsilon}.
\end{equation}

\subsection{Gaussian-Aware Information Fusion}
\label{subsec:fusion}

\begin{table*}[t]
\centering
\caption{Tracking performance (ATE RMSE ↓ [cm]) and Runtime Performance (FPS) on TUM RGB-D No Structure/No Texture Scenes. \textbf{Bold} and \underline{underline} indicate the best and second-best results, respectively.}
\label{tab:ate_comparison_main}
\setlength{\tabcolsep}{5pt} 
\resizebox{\textwidth}{!}{
\begin{tabular}{l|l|cc|cccc|cccc|cc}
\toprule
& & & \multirow{2}{*}{\shortstack[b]{\textbf{\%} \\ \textbf{tracked}}} & \multicolumn{4}{c|}{\textbf{No Structure}} & \multicolumn{4}{c|}{\textbf{No Texture}} & & \\
& \textbf{Method} & \textbf{Tracker} & & \multicolumn{4}{c|}{ATE RMSE (cm) $\downarrow$} & \multicolumn{4}{c|}{ATE RMSE (cm) $\downarrow$} & \textbf{FPS} & \textbf{BA} \\ \cmidrule(lr){5-8} \cmidrule(lr){9-12}
& & & & \textbf{near} & \textbf{near\_val} & \textbf{far} & \textbf{far\_val} & \textbf{near} & \textbf{near\_val} & \textbf{far} & \textbf{far\_val} & & \\ \midrule
\multirow{2}{*}{\rotatebox[origin=c]{90}{\shortstack{Not\\RT}}} & SplaTAM \cite{keetha_splatam_2024} & Photometric & \textbf{100} & \underline{4.49} & \underline{3.65} & 9.49 & 5.32 & \underline{2.27} & 2.81 & \underline{4.54} & \underline{4.14} & 0.54 & \checkmark \\
& MonoGS \cite{matsuki_gaussian_2024} & Photometric & \textbf{100} & 11.86 & 5.37 & \textbf{4.10} & \textbf{3.68} & 26.91 & 28.74 & 10.39 & 18.63 & 1.89 & \checkmark \\ \midrule
\multirow{4}{*}{\rotatebox[origin=c]{90}{\shortstack{\textbf{Real-}\\\textbf{Time}}}} & Photo-SLAM \cite{huang_photo-slam_2024} & ORB & \underline{68.4} & \textbf{2.27} & \textbf{3.31} & \underline{5.36} & \underline{3.83} & X & X & X & X & 31.48 & \checkmark \\
& GS-ICP SLAM \cite{leonardis_rgbd_2025} & ICP & 41.0 & 194.93 & 197.65 & 116.16 & 143.16 & \textbf{1.55} & \underline{1.10} & 6.07 & 4.53 & \textbf{67.90} & -- \\ \cmidrule(lr){2-14}
& \textbf{Ours} & ICP+ORB & \textbf{100} & 7.79 & 9.66 & 16.85 & 12.96 & \textbf{1.55} & \textbf{1.09} & \textbf{2.19} & \textbf{3.89} & \underline{40.14} & -- \\ \bottomrule
\end{tabular}
}
\end{table*}

To stabilize the optimization process when there is geometric degeneracy, specifically where the lateral information remains small, we use the complementary strengths of visual feature-based matching to define a low-cost frame-to-frame prior estimation: 

\begin{equation}
    \mathbf{T}^*_{\text{prior}} = \arg \min_{\mathbf{T} \in \mathrm{SE}(3)} \sum_{i \in \mathcal{X}} \rho \left( \left\| \mathbf{u}_i - \pi(\mathbf{T} \cdot \mathbf{P}_i) \right\|^2_{\bm{\Sigma}_i} \right)
\end{equation}

Then, for each degenerate axis $k$, we perform anisotropic information injection with the prior into the degenerate subspace containing low information $\Lambda$: 
\begin{equation}
    \mathbf{H} =  \mathbf{H}_{\text{ICP}} + \sum_{k \in \mathcal{K}} w_k \cdot \left(  \mathbf{v}_k^\top {\mathbf{\Omega}}_{\text{prior}} \mathbf{v}_k \right) \mathbf{v}_k \mathbf{v}_k^\top
\end{equation}

where $w_k = \exp\left(-\alpha (1 - \eta)\right) \in [0,1]$ is a soft gating score to control the relative strength of the prior added. If the planarity ratio is too high, we directly add the entire prior. We have therefore fused compensatory information directly into the ICP optimizer without conflicting with the mapping objective.

\section{EXPERIMENTS}

\subsection{Implementation and Experiment Setup}
\textbf{Datasets and Metrics. }We utilize the TUM RGB-D \cite{sturm_benchmark_2012} and Replica \cite{straub_replica_2019} dataset to evaluate the performance of our method. We evaluate against a variety of sequences with different levels of visual textures, geometry etc. The metric used for trajectory evaluation is the root mean square error (RMSE) absolute trajectory error (ATE). We evaluate map quality using the standard measures of PSNR (dB), LPIPS \cite{zhang_unreasonable_2018}, and SSIM \cite{zhou_wang_image_2004}. Lastly, we provide a qualitative visual comparison of the map and trajectory quality. We provide additional visualization of our method in Section~\ref{subsec:appx_qual_deg}.

\textbf{Implementation Details. } All of our evaluations were performed on a computer equipped with an i9-13900K CPU and an RTX 4090 GPU. We implemented all the comparisons for GS-ICP SLAM \cite{leonardis_rgbd_2025}, Photo-SLAM \cite{huang_photo-slam_2024}, MonoGS \cite{matsuki_gaussian_2024} and SplaTAM \cite{keetha_splatam_2024} using the open-source implementations with the default configurations, unless the results were explicitly provided. For partial or otherwise incomplete runs due to failed initialization (e.g., lack of features) or lost tracking, we denote by ``X". ``BA" refers to the presence of Bundle Adjustment or other forms of window-based pose adjustment. Frame-per-second (FPS) is defined for the entire system, and ``\% tracked" refers to the number of valid poses predicted along the trajectory not exceeding 1.0m. We empirically fix the hyperparameter $\tau_{\text{deg}} = 200$ for all experiments, demonstrating that our method does not require environment-specific tuning.

\subsection{Tracking Performance Evaluation}
In Table~\ref{tab:ate_comparison_main}, we show that our method is the only real-time 3DGS-SLAM capable of robust survival in geometrically and photometrically degenerate scenes. We attain state-of-the-art performance on `No Texture' scenes, whereas other feature-based methods experience tracking failure.

\begin{table*}[t]
\centering
\begin{threeparttable} 
\caption{Tracking performance (ATE RMSE ↓ [cm]) on Replica Scenes. \textbf{Bold} and \underline{underline} indicate the best and second-best results, respectively.}
\label{tab:replica_comp}
\begin{tabular}{l|l|cccccccc|c}
\toprule
\textbf{Type} & \textbf{Method} & \textbf{R0} & \textbf{R1} & \textbf{R2} & \textbf{O0} & \textbf{O1} & \textbf{O2} & \textbf{O3} & \textbf{O4} & \textbf{Avg.} \\ \midrule
\multirow{4}{*}{\rotatebox[origin=c]{90}{\shortstack{Not\\RT}}} 

& SplaTAM \cite{keetha_splatam_2024} & \underline{0.29} & \underline{0.35} & 0.28 & 0.49 & \underline{0.21} & 0.31 & 0.34 & 0.57 & 0.36 \\

& GS-SLAM \cite{yan_gs-slam_2024}& 0.48 & 0.53 & 0.33 & 0.52 & 0.41 & 0.59 & 0.46 & 0.70 & 0.50 \\ 
& MonoGS \cite{matsuki_gaussian_2024} & 0.48 & 0.36 & 0.34 & \underline{0.44} & 0.52 & \underline{0.23} & \textbf{0.16} & 2.53 & 0.58 \\ 
& Point-SLAM \cite{sandstrom_point-slam_2023} & 0.59 & 0.51 & 0.32 & 0.45 & 0.46 & 0.48 & 0.61 & 0.87 & 0.54 \\ \midrule

\multirow{3}{*}{\rotatebox[origin=c]{90}{\shortstack{\textbf{Real-}\\\textbf{Time}}}} 

& Photo-SLAM \cite{huang_photo-slam_2024} & - & - & - & - & - & - & - & - & 0.60 \\

& GS-ICP SLAM* \cite{leonardis_rgbd_2025} & \textbf{0.15} & \textbf{0.16} & \underline{0.11} & \textbf{0.19} & \textbf{0.12} & \textbf{0.16} & \underline{0.18} & \underline{0.21} & \textbf{0.16} \\

& \textbf{Ours} & \textbf{0.15} & \textbf{0.16} & \textbf{0.10} & \textbf{0.19} & \textbf{0.12} & 0.57 & 0.47 & \textbf{0.19} & \underline{0.24} \\ \bottomrule
\end{tabular}%
\begin{tablenotes}[flushleft] 
    \small
    \item * indicates reproduced results.
\end{tablenotes}
\end{threeparttable}
\end{table*}
\begin{figure}[h]
    \centering
    \includegraphics[width=0.95 \columnwidth]{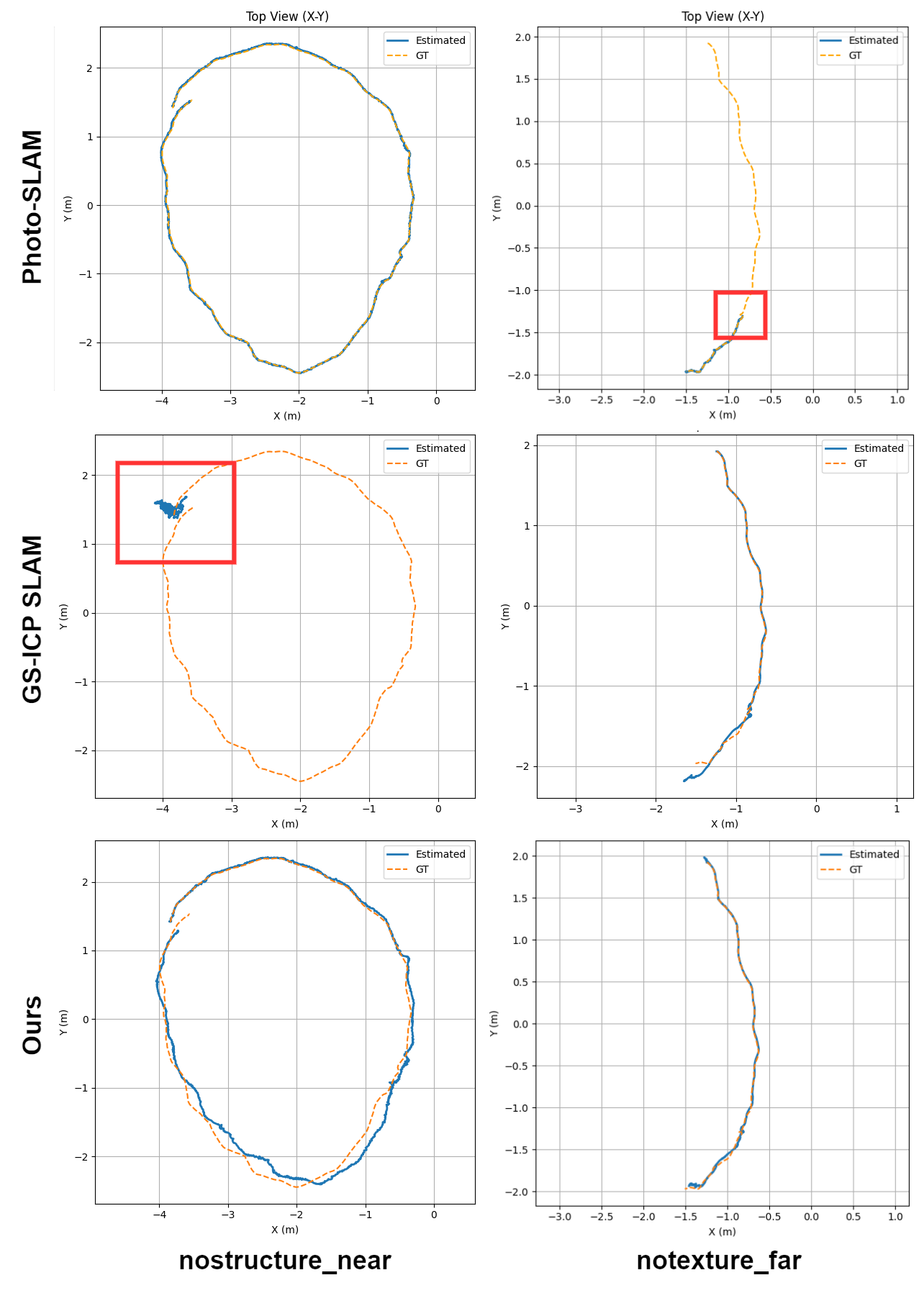}
    \caption{Trajectory plot (top view) of real-time methods for structureless (left) and featureless (right) scenes. GS-ICP SLAM fails to track the structureless case, while Photo-SLAM loses tracking in the featureless case, ours successfully tracks both.}
    \label{fig:traj}
\end{figure}

Although our method yields a higher tracking error in the `No Structure' scenes compared to feature-based methods like Photo-SLAM, this represents a deliberate trade-off for robustness. `No Structure' scenes exhibit rich photometric gradients but lack geometric constraints, allowing pure visual feature trackers to achieve high local precision. However, aggressively optimizing for these specific conditions creates brittle architectures susceptible to collapse under ambiguous conditions. By applying a consistent spectral regularization constraint across all environments, we explicitly trade marginal local accuracy for absolute global robustness (see Fig.~\ref{fig:traj}). We provide details of the exact outlier condition for this scene in Section~\ref{subsec:appx_dual_fail}.

Crucially, this resilience requires no environment-specific tuning. As shown in Table~\ref{tab:replica_comp} (Replica), we evaluate our system without adjusting any detection parameters from Table~\ref{tab:ate_comparison_main}. Under these synthetic conditions, our algorithm correctly identifies the structural stability, leaving the geometric prior inactive. The marginal error accumulated in specific scenes (e.g., O2, O3) is caused by brief threshold crossing of our degeneracy detection algorithm, since we did not perform scene-specific hyperparameter tuning.

Furthermore, in standard real-world environments (see Table~\ref{tab:standard_scenes_comp}), our method maintains competitive generalization. In such environments with fluctuating geometric stability (e.g., fr3\_office), the prior is frequently switched, inducing minor drift. This explicitly confirms our system's ability to maintain continuous tracking survival in unpredictable conditions.

\begin{table}[t]
\centering
\caption{Tracking performance (ATE RMSE ↓ [cm]) on standard TUM RGB-D Scenes. \textbf{Bold} and \underline{underline} indicate the best and second-best results, respectively.}
\label{tab:standard_scenes_comp}
\begin{tabular}{l|l|ccc}
\toprule
\textbf{Type} & \textbf{Method} & \textbf{fr1/desk} & \textbf{fr2/xyz} & \textbf{fr3/office} \\ \midrule
\multirow{4}{*}{\rotatebox[origin=c]{90}{\shortstack{\textbf{Not}\\\textbf{Real-Time}}}}
& SplaTAM \cite{keetha_splatam_2024} & 3.35 & \underline{1.24} & 5.16 \\
& MonoGS \cite{matsuki_gaussian_2024}  & \textbf{1.48} & 1.45 & \underline{1.50} \\ 
& GS-SLAM \cite{yan_gs-slam_2024} & 3.30 & 1.30 & 6.60 \\ 
& Point-SLAM \cite{sandstrom_point-slam_2023} & 4.34 & 1.31 & 3.48 \\ \midrule
\multirow{4}{*}{\rotatebox[origin=c]{90}{\shortstack{\textbf{Real-}\\\textbf{Time}}}}
& Photo-SLAM \cite{huang_photo-slam_2024} & \underline{2.60} & \textbf{0.35} & \textbf{1.00} \\
& GS-ICP SLAM \cite{leonardis_rgbd_2025} & 3.07 & 1.79 & 2.46 \\
& \textbf{Ours} & 3.05 & 1.75 & 3.82 \\
\bottomrule
\end{tabular}%
\end{table}

\begin{figure}[h]
    \centering
    \includegraphics[width=\columnwidth]{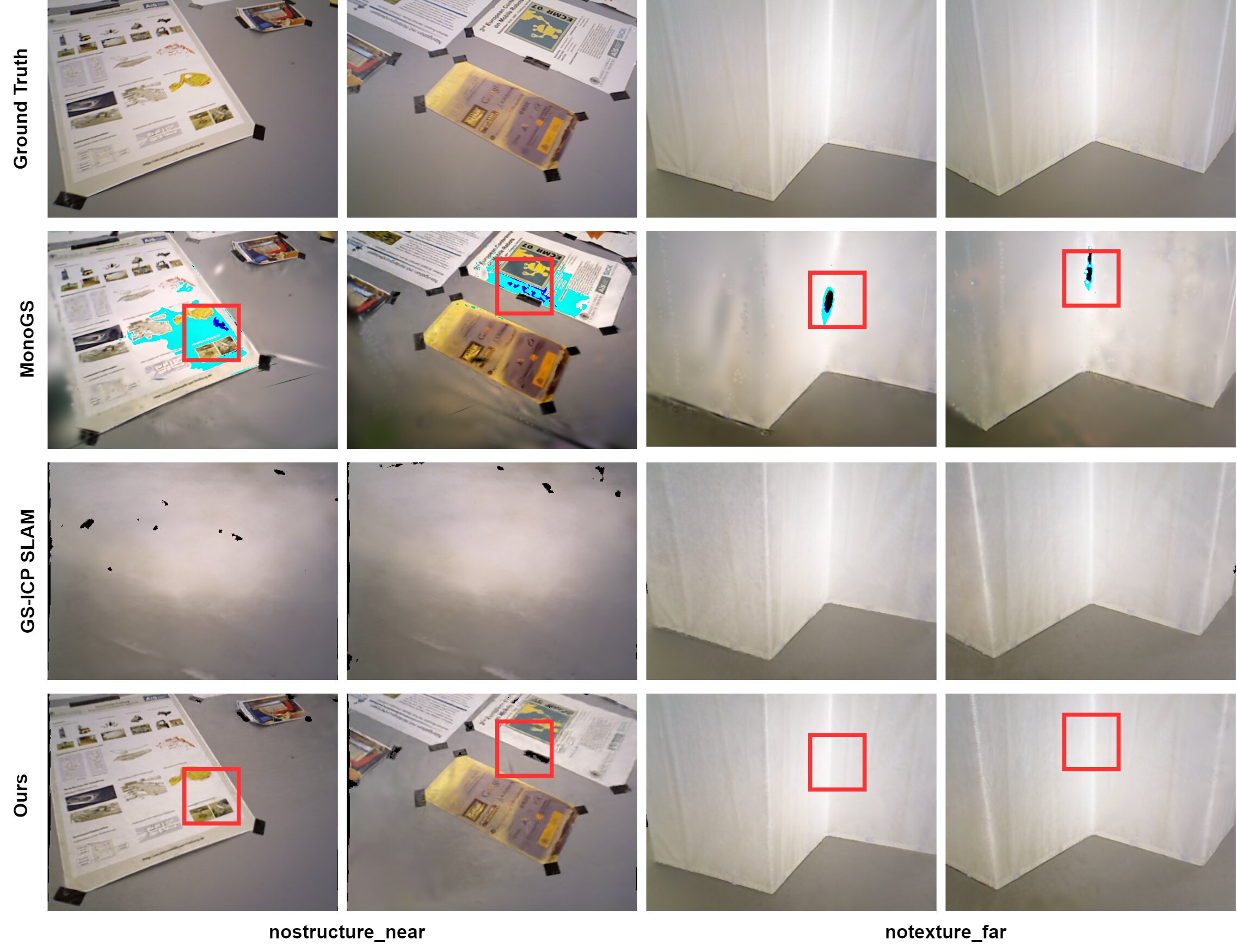}
    \caption{Qualitative Scene Rendering Performance on TUM RGB-D No Structure/No Texture Scenes. Ours produces high fidelity renders, while other works produce color-distorted images due to noisy tracking as indicated in \textcolor{red}{red} (MonoGS), or fail to render the scene due to failed tracking (GS-ICP SLAM). }
    \label{fig:renders}
\end{figure}

\begin{table}[H]
\centering
\caption{Evaluation of 3D Reconstruction Quality on TUM RGB-D No Structure/No Texture Scenes. \textbf{Bold} and \underline{underline} indicate the best and second-best results, respectively.}
\label{tab:rendering_quality}
\begin{tabular}{l|l|cccc}
\toprule
\multirow{2}{*}{\textbf{Method}} & \multirow{2}{*}{\textbf{Metric}} & \multicolumn{2}{c}{\textbf{No Structure}} & \multicolumn{2}{c}{\textbf{No Texture}} \\
& & \textbf{near} & \textbf{far} & \textbf{near} & \textbf{far} \\ \midrule

\multirow{3}{*}{MonoGS \cite{matsuki_gaussian_2024}} 
& PSNR $\uparrow$  & 20.97 & \textbf{26.59} & 20.82 & 21.95 \\
& SSIM $\uparrow$  & 0.811 & 0.863 & 0.895 & 0.902 \\
& LPIPS $\downarrow$ & 0.337 & 0.220 & 0.447 & 0.363 \\ \midrule

\multirow{3}{*}{SplaTAM \cite{keetha_splatam_2024}} 
& PSNR $\uparrow$  & 20.71 & 21.36 & \underline{26.71} & \textbf{30.70} \\
& SSIM $\uparrow$  & \textbf{0.827} & \textbf{0.903} & 0.924 & \textbf{0.949} \\
& LPIPS $\downarrow$ & \textbf{0.249} & \textbf{0.124} & 0.184 & \underline{0.136} \\ \midrule

\multirow{3}{*}{GS-ICP SLAM \cite{leonardis_rgbd_2025}} 
& PSNR $\uparrow$  & 18.02 & 18.17 & 27.97 & 28.15 \\
& SSIM $\uparrow$  & 0.741 & 0.750 & \underline{0.931} & 0.929 \\
& LPIPS $\downarrow$ & 0.507 & 0.452 & \textbf{0.135} & 0.137 \\ \midrule

\multirow{3}{*}{\textbf{Ours}} 
& PSNR $\uparrow$  & \textbf{22.61} & \underline{24.62} & \textbf{29.19} & \underline{30.32} \\
& SSIM $\uparrow$  & \underline{0.768} & \underline{0.867} & \textbf{0.948} & \underline{0.942} \\
& LPIPS $\downarrow$ & \underline{0.260} & \underline{0.163} & \underline{0.138} & \textbf{0.127} \\
\bottomrule
\end{tabular}%
\end{table}

\subsection{Mapping Performance Evaluation}
We show that our real-time method maintains high mapping fidelity across all of the methods on the challenging scenes as shown in Table~\ref{tab:rendering_quality}. Our method performs competitively with other works, including much heavier systems, because our solution explicitly decouples the degeneracy correction of the tracker and the Gaussian primitives for mapping.
We provide qualitative examples in Fig.~\ref{fig:renders}, showing that our method represents subtle colors, textures and details well. We omit works where the tracking is unsuccessful (e.g., Photo-SLAM).

\subsection{Qualitative Evaluation of Degeneracy Detection Mechanism}
\label{subsec:appx_qual_deg}

\begin{figure}[h]
    \centering
    \includegraphics[width=0.92\columnwidth]{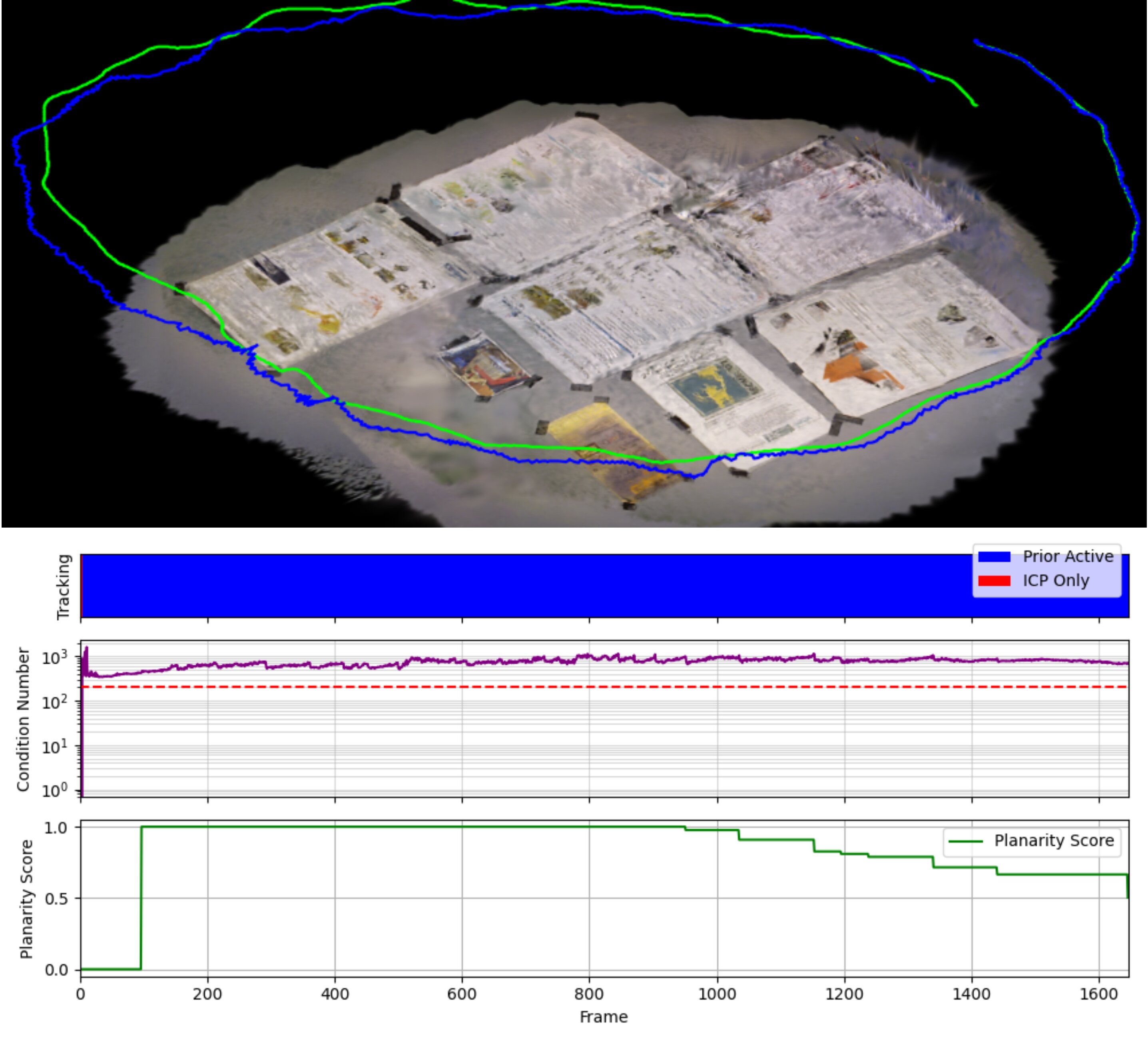}
    \caption{Degeneracy detection visualization on the nostructure\_near.}
    \label{fig:nostr}
\end{figure}

\begin{figure}[h]
    \centering
    \includegraphics[width=0.92\columnwidth]{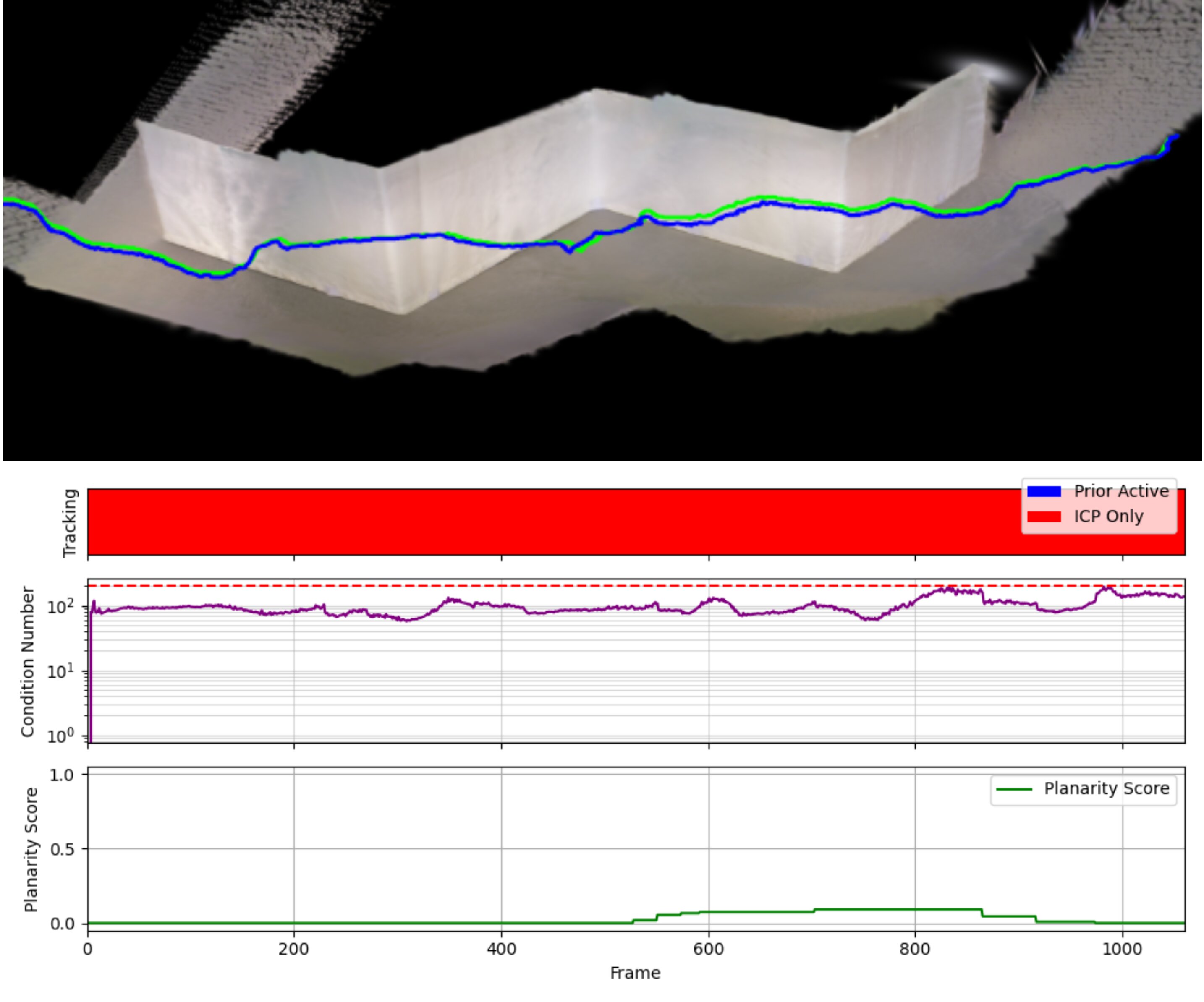}
    \caption{Degeneracy detection visualization on the notexture\_near.}
    \label{fig:notex}
\end{figure}

\begin{figure}[h]
    \centering
    \includegraphics[width=0.92\columnwidth]{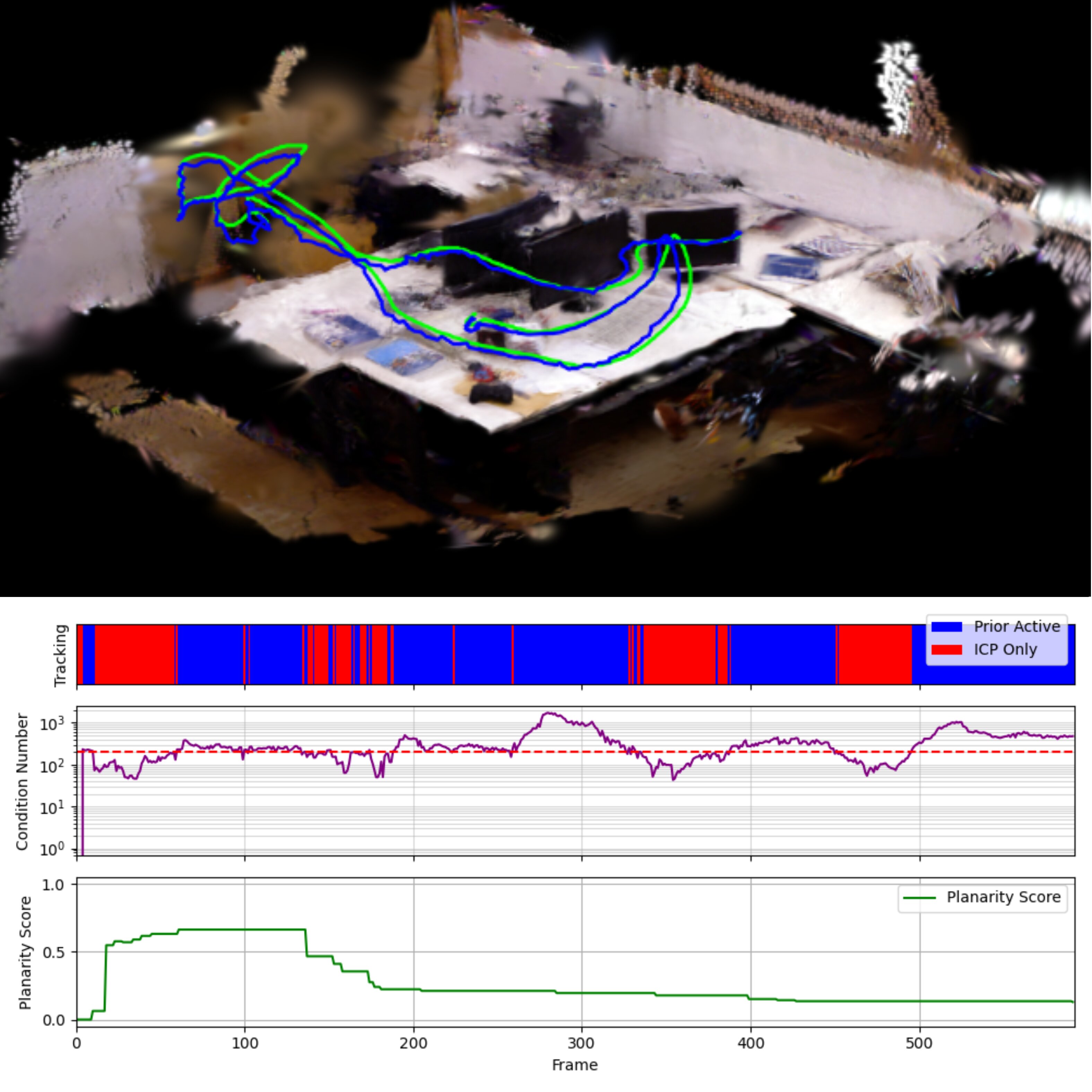}
    \caption{Degeneracy detection visualization on fr1\_desk.}
    \label{fig:norml}
\end{figure}

We provide qualitative visualizations to investigate our degeneracy detection mechanism in both structureless and textureless scenes (see Fig.~\ref{fig:nostr}-\ref{fig:notex}). We also analyze a scene without those specific extreme cases (see Fig.~\ref{fig:norml}), which we demonstrated previously can \textit{benefit} from our method (see Table~\ref{tab:standard_scenes_comp}), implying that our prior information fusion provides helpful constraints in these non-degenerate cases.

``Planarity Score" refers to the ratio $\eta$ in Section~\ref{subsec:awareness}, while the row denoted by ``Tracking" is a binary indicator of whether the prior is active based on the eigenvalue measure.

For the structureless and textureless scenes, we show that our mechanism stably detects the respective condition over the entire trajectory. Additionally, the planarity score accurately describes the scene's dominant normal covariance, experimentally supporting our method.

In the non-degenerate scene illustrated in Fig.~\ref{fig:norml}, we show that the condition number threshold is repeatedly crossed many times, but the tracker is still able to perform well, and in some scenes outperform the baseline GS-ICP SLAM. This demonstrates that our system is not brittle under more ambiguous or borderline degenerate conditions.

\begin{figure}[h]
    \centering
    \includegraphics[width=\columnwidth]{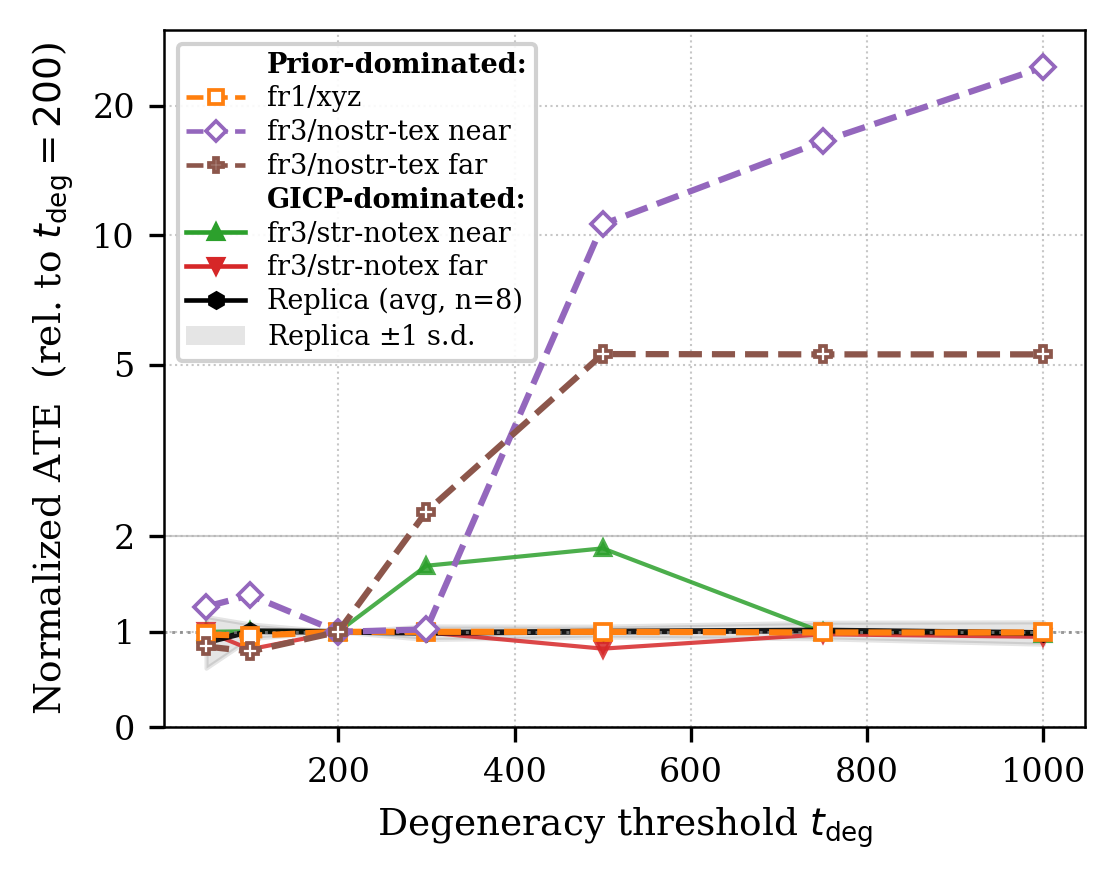}
    \caption{Per-sequence ATE vs. degeneracy threshold $\tau_{\text{deg}}$, with each sequence normalized to its own ATE at $\tau_{\text{deg}}{=}200$.}
    \label{fig:sweep}
\end{figure}

We additionally perform a hyperparameter sweep of $\tau_{\text{deg}}$ over a wide range in Fig.~\ref{fig:sweep}. In those scenes reliant on the prior (e.g., no structure), ATE increases as the prior is triggered less with higher thresholds. Scenes that do not use the prior are largely unaffected. In particular, failures occur where the threshold is too high for our mechanism to trigger, not when the threshold is set too low.

\subsection{Analysis of Dual-Degeneracy Cases}
\label{subsec:appx_dual_fail}
\begin{figure}[h]
    \centering
    \includegraphics[width=\columnwidth]{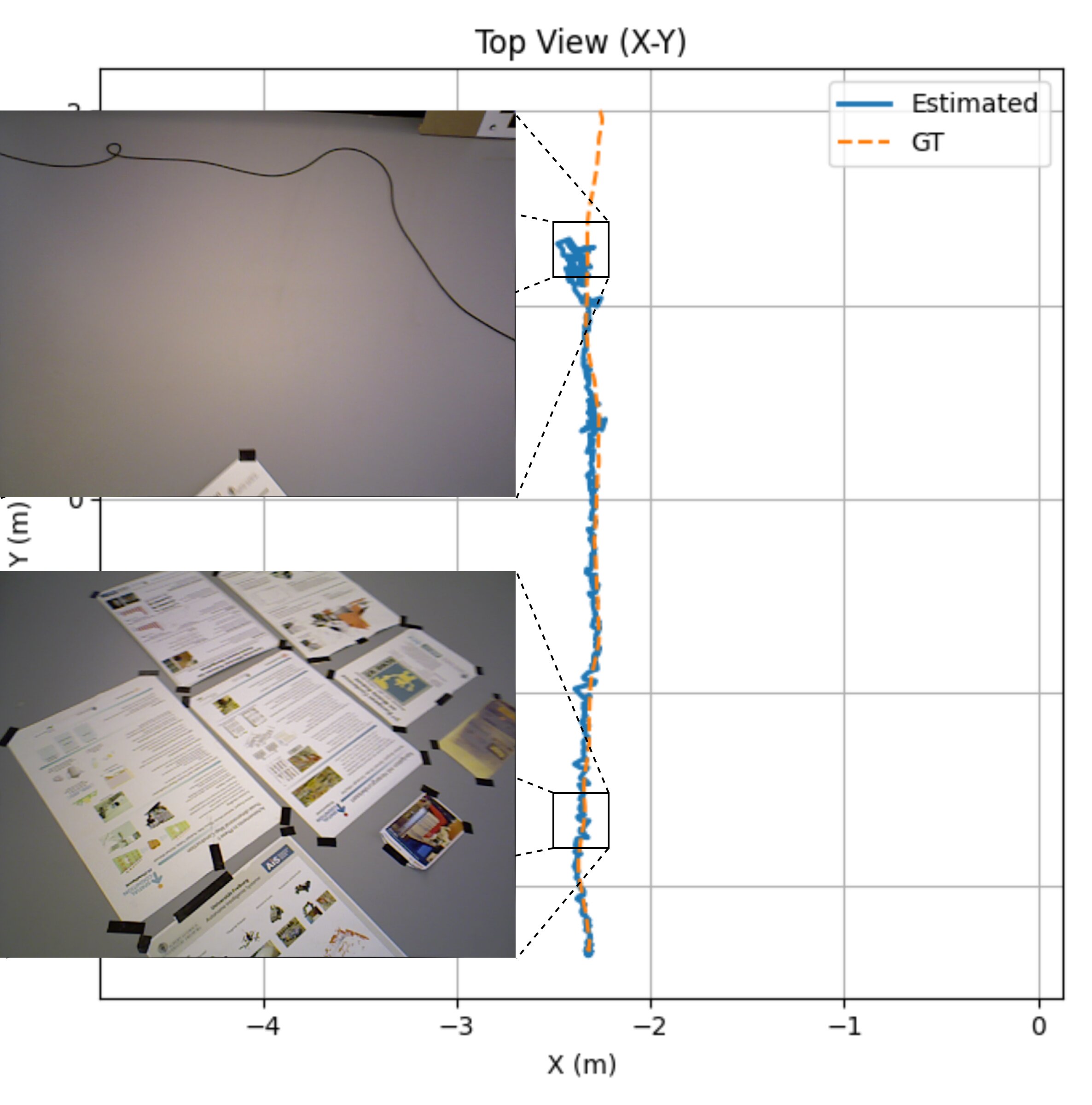}
    \caption{Trajectory plot and visualization on nostructure\_far scene.}
    \label{fig:failcase}
\end{figure}

In this section, we briefly analyze the case where a scene is \textit{both} featureless and structureless. We show in Fig.~\ref{fig:failcase} that near the end of the trajectory, the nostructure far/far\_val scenes contain both a case of extreme lack of features and near complete planarity despite the data sequence's test objective. In this specific corner case, the lightweight frame-to-frame prior we used lacks feature correspondences and is thus rejected and set to zero motion (instead of accepting a wrong RANSAC estimate) for robustness. Therefore, our proposed information fusion module correctly locks the pose in place due to an overall lack of constraints from both sources.

\subsection{Ablation Study}
In this section, we perform various experiments to verify the effectiveness of our method. We list the ablation studies conducted in Table~\ref{tab:ablation}. The ``-" represents subtraction of various components of our method.

The first row represents the performance reported for our entire system. ``- Soft Gating" refers to the removal of the soft gating mechanism $w_i$, replacing it with binary inclusion of the prior elements. ``- Planarity score" refers to the removal of the planarity-based detection, relying only on the eigenvalue-based measure. In this case, the insertion of the prior is equivalent to a hard switch, contingent only on the eigenvalues of each axis and without explicitly considering the scene's spatial properties.

Lastly, we set the minimum scale $s_{\perp} \approx k \cdot \sigma_z(d)$ where $\sigma_z(d)$ is the depth noise, to test whether forcing the mapper to not overflatten the Gaussian can improve tracking. We report the results as ``Scale" and find that it worsens both; the optimizer competes with the tracker and spawns more Gaussians to compensate, causing the correspondence quality to worsen.

\begin{table}[h]
\centering
\caption{Ablation Study of System Components: Tracking Performance (ATE RMSE ↓ [cm]) on TUM RGB-D No Structure/No Texture Scenes.}
\label{tab:ablation}
\begin{tabular}{l|cc|cc}
\toprule
\multirow{2}{*}{\textbf{Configuration}} & \multicolumn{2}{c|}{\textbf{No Structure}} & \multicolumn{2}{c}{\textbf{No Texture}} \\
& \textbf{near} & \textbf{far} & \textbf{near} & \textbf{far} \\ \midrule
\textbf{Ours} & \textbf{7.79} & \textbf{16.85} & \textbf{1.55} & \textbf{2.19} \\ 
- Soft gating & 7.91 & 22.43 & 1.88 & 2.52 \\
- Planarity score & 18.49 & 75.78 & 1.63 & 2.37 \\ \midrule
Scale & 8.24 & 22.64 & 3.58 & 2.57  \\ 
\bottomrule
\end{tabular}%
\end{table}

We also ablate in Table~\ref{tab:ablation_prior} the necessity of our feature-based prior. ``Prior only" refers to directly using the prior pose estimate, while for ``prior $\leftarrow$ CVM", we replace the prior with a Constant Velocity Model (CVM), showing that this problem cannot be trivially solved with a simple regularizer.

\begin{table}[h]
\centering
\caption{Ablation Study of Prior Injection and Regularization: Tracking Performance (ATE RMSE ↓ [cm]) on TUM RGB-D No Structure/No Texture Scenes.}
\label{tab:ablation_prior}
\begin{tabular}{l|cc|cc}
\toprule
\multirow{2}{*}{\textbf{Configuration}} & \multicolumn{2}{c|}{\textbf{No Structure}} & \multicolumn{2}{c}{\textbf{No Texture}} \\
& \textbf{near} & \textbf{far} & \textbf{near} & \textbf{far} \\ \midrule
\textbf{Ours} & \textbf{7.79} & \textbf{16.85} & \textbf{1.55} & \textbf{2.19} \\ 
Prior only & 6.77 & 45.29 & 103.67 & 21.18 \\
Prior $\leftarrow$ CVM & 131.18 & 55.87 & 1.68 & 2.43 \\ 
\bottomrule
\end{tabular}%
\end{table}

These results confirm that each proposed component of our method contributes meaningfully to the results, and our observation-based planarity ratio complements the Hessian eigenvalue analysis in reusing the Gaussian information to improve robustness against spurious prior injection.
\section{LIMITATIONS}
In this work, we have investigated and addressed the problem of a scene being \textit{either} featureless or textureless. However, as detailed in Section~\ref{subsec:appx_dual_fail}, in purely dual-degenerate scenarios, our system is able to maintain safe pose lock, but reaches an unobservable state due to a fundamental lack of visual or geometric data. Future work may consider incorporating low-cost IMU sensors to provide a motion estimate should tracking through these conditions be required.

\section{CONCLUSIONS}

We propose Spectral GS-SLAM, which addresses the problem of failed tracking and the lack of robustness of existing lightweight 3DGS-SLAM methods in transiently degenerate environments. Ensuring that the SLAM system is robust in such conditions ensures reliable long-term operation. We achieve this while maintaining high map quality and generalization to non-challenging or ambiguous scenes, showing that our proposed fusion strategy does not require environment-specific hyperparameter tuning.

Our method demonstrates that we can leverage the strengths of the explicit Gaussian primitives to compute a planarity ratio for scene planarity understanding with negligible overhead, and directly stabilize the tracking optimization with a low-cost visual feature prior, without conflicting with the mapping objective. Overall, these results suggest our system's potential applicability in diverse real-world environments.


\bibliographystyle{IEEEtran} 
\bibliography{references}    

\addtolength{\textheight}{-12cm}   
\end{document}